\documentclass{article}

\PassOptionsToPackage{numbers, compress}{natbib}

\usepackage[preprint]{neurips_2024}

\usepackage[usenames,dvipsnames,svgnames,table]{xcolor}
\usepackage[colorlinks,allcolors=Blue]{hyperref}       
\usepackage{url}            
\usepackage{booktabs}       
\usepackage{amsfonts}       

\usepackage{nicefrac}       
\usepackage[activate=true,
final,
tracking=false,   
kerning=true,
spacing=true,
factor=1050,
stretch=30,
shrink=30]{microtype}      

\usepackage{amsmath,amsfonts,bm}

\def\eqref#1{equation~\ref{#1}}

\def\1{\bm{1}}

\DeclareMathAlphabet{\mathsfit}{\encodingdefault}{\sfdefault}{m}{sl}
\SetMathAlphabet{\mathsfit}{bold}{\encodingdefault}{\sfdefault}{bx}{n}

\newcommand{\E}{\mathbb{E}}

\DeclareMathOperator*{\argmax}{arg\,max}
\DeclareMathOperator*{\argmin}{arg\,min}

\usepackage[utf8]{inputenc} 
\usepackage[T1]{fontenc}    
\usepackage{url}            
\usepackage{booktabs}       
\usepackage{amsfonts}       
\usepackage{nicefrac}       
\usepackage{microtype}      
\usepackage{xcolor}         
\usepackage{amssymb}
\usepackage[ruled]{algorithm2e}
\usepackage{subfigure}
\usepackage{amsmath}
\usepackage{amsthm}
\usepackage{adjustbox}
\usepackage{multirow}
\usepackage{booktabs}
\usepackage{amsmath}
\usepackage{svg}
\usepackage{arydshln}
\usepackage{soul}
\usepackage{cleveref}
\usepackage{wrapfig}
\usepackage{macros}
\usepackage{enumitem}
\usepackage{arydshln}
\usepackage{xcolor}
\usepackage{amsthm}
\usepackage{thmtools}
\usepackage{bbm}
\usepackage{stmaryrd}
\crefformat{section}{\textcolor{blue}{\S#2#1#3}}
\crefformat{subsection}{\textcolor{blue}{\S#2#1#3}}
\crefformat{subsubsection}{\textcolor{blue}{\S#2#1#3}}
\crefformat{equation}{\textcolor{orange}{Eq.~(#1)}}

\definecolor{mydarkblue}{rgb}{0,0.08,0.45}
\theoremstyle{plain}

\definecolor{seamcolor}{HTML}{5758BB}
\definecolor{White}{rgb}{1, 1, 1}
\definecolor{Periwinkle}{rgb}{0, 0, 0}
\definecolor{myblue}{rgb}{0.82, 0.94, 0.75}
\definecolor{mygreen}{rgb}{0.64, 0.76, 0.68}
\definecolor{myyellow}{rgb}{0.88, 0.54, 0.35}
\definecolor{mygreen}{rgb}{0.68, 0.9, 0.8}
\definecolor{mypink}{rgb}{0.2, 0.87, 0.2}

\newcommand{\contrast}{\textsc{Contrast Instructions}}
\newcommand{\seam}{{\textcolor{seamcolor}{\textsc{SEAM}}}}
\usepackage[most]{tcolorbox}

\colorlet{LightGray}{White!98!Periwinkle}

\declaretheoremstyle[
    name=Definition,
]{thmsty}
\declaretheorem[style=thmsty]{definition}

\tcolorboxenvironment{definition}{enhanced jigsaw,colback=LightGray,drop shadow,boxrule=0.9pt, boxsep=0.1pt,left=4pt,right=4pt,top=4pt,bottom=4pt}

\declaretheoremstyle[
    name=Prompt,
]{thmsty}
\declaretheorem[style=thmsty]{prompt}
\tcolorboxenvironment{prompt}{enhanced jigsaw,colback=LightGray,drop shadow,boxrule=0.9pt, boxsep=0.1pt,left=4pt,right=4pt,top=4pt,bottom=4pt}

\title{It Takes Two: On the Seamlessness between Reward and Policy Model in RLHF}

\author{%
  Taiming Lu$^\clubsuit$ \\
  JHU\\
  \And
  Lingfeng Shen$^\clubsuit$\\
  Bytedance \\
  \And
  Xinyu Yang$^\clubsuit$\\
  CMU \\
  \And
  Weiting Tan \\
  JHU \\
  \AND
  Beidi Chen \\
  CMU \\
  \And
  Huaxiu Yao \\
  UNC-Chapel Hill \\
}

\begin{document}

\maketitle

\blfootnote{$\clubsuit$ indicates equal contributions.}

\begin{abstract}

    Reinforcement Learning from Human Feedback (RLHF) involves training policy models (PMs) and reward models (RMs) to align language models with human preferences. Instead of focusing solely on PMs and RMs independently, we propose to examine their interactions during fine-tuning, introducing the concept of \textbf{seamlessness}. Our study starts with observing the saturation phenomenon, where continual improvements in RM and PM do not translate into RLHF progress. Our analysis shows that RMs fail to assign proper scores to PM responses, resulting in a 35\% mismatch rate with human preferences, highlighting a significant discrepancy between PM and RM. To measure seamlessness between PM and RM without human effort, we propose an automatic metric, \seam{}. \seam{} quantifies the discrepancies between PM and RM judgments induced by data samples. We validate the effectiveness of \seam{} in data selection and model augmentation. Our experiments demonstrate that (1) using \seam{}-filtered data for RL training improves RLHF performance by 4.5\%, and (2) \seam{}-guided model augmentation results in a 4\% performance improvement over standard augmentation methods. The code is accessible {here}: \url{https://github.com/TaiMingLu/seamless}

\end{abstract}

\section{Introduction}
Reinforcement learning from human feedback (RLHF) has emerged as a popular technique to optimize and align a language model with human preferences \citep{stiennon2020learning,nakano2021webgpt,menick2022teaching,glaese2022improving,ouyang2022training,touvron2023llama,achiam2023gpt,bai2023qwen,rafailov2024direct}.
RLHF provides a natural solution for optimizing non-differentiable, scalar objectives for language models and has been the centerpiece of recent state-of-the-art large language models (LLMs) \citep{lu2022quark,hejna2023few,go2023aligning,korbak2023pretraining,achiam2023gpt,openai2023gpt4}. In RLHF, a \emph{reward model} (RM) generates scalar rewards for a \emph{policy model} (PM) generated outputs as supervision signals during reinforcement learning. 
Since policy gradient methods \citep{schulman2017proximal} optimize based on such signal, the PM and RM inevitably dictate the behavior of the resultant RLHF model.
{As such, the properties of RMs (or PMs) and their impact on RLHF models have become points of interest for the community \citep{gao2023scaling,zhu2023principled,dong2023raft, gao2023scaling, shen2023trickle}.} Unlike prior work that examines the individual capabilities of each model, in this work, we introduce and explore the concept of \emph{seamlessness} between the PM and RM, focusing on their interactions.


\begin{figure}
\vspace{-1em}
\centering
    \includegraphics[width=1.0\textwidth]{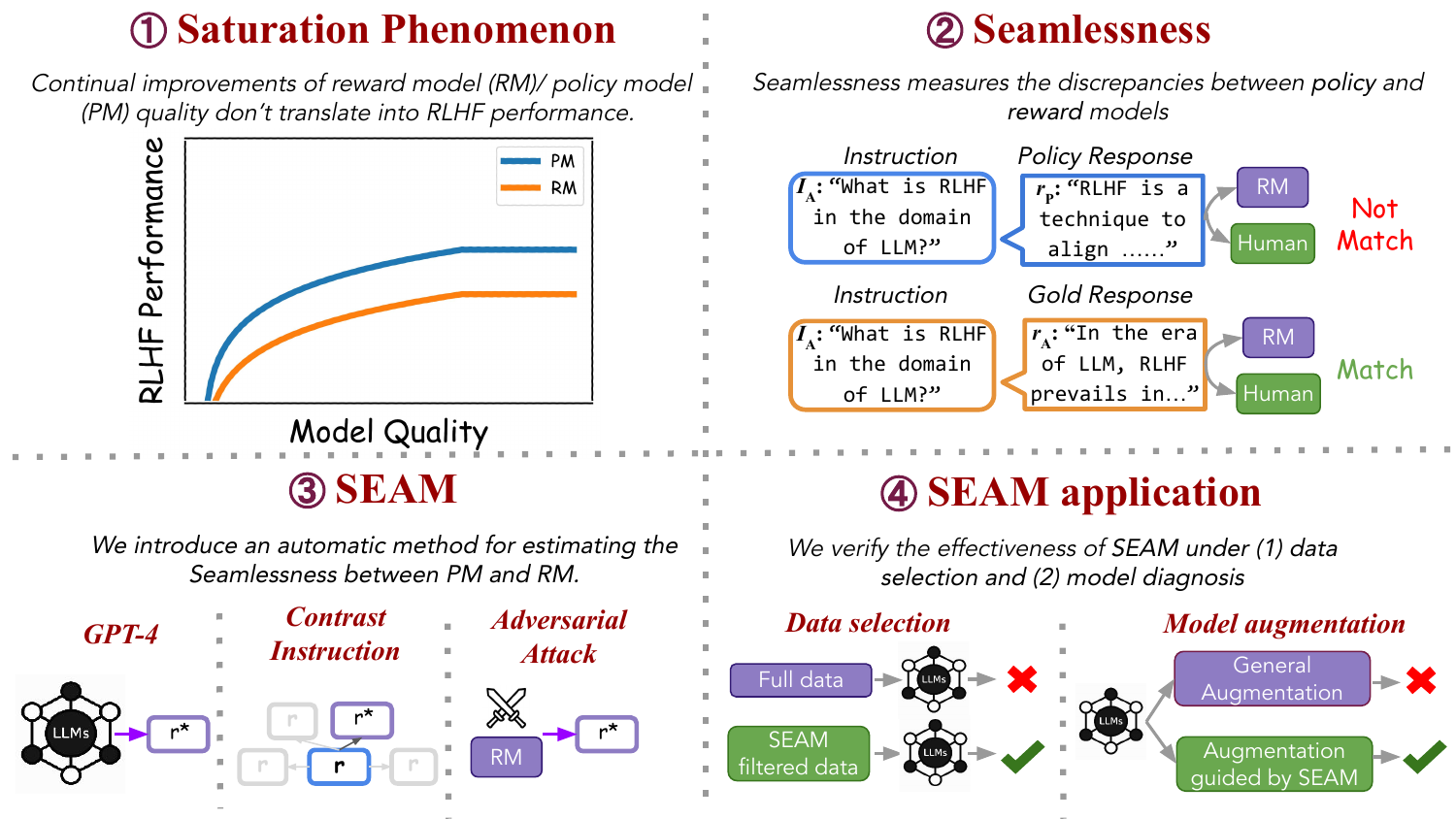}
\vspace{-1em}
\caption{We introduce the concept of \textit{Seamlessness} to measure the \emph{discrepancies} between reward and policy models as supported by human evaluation. To automate measuring the \textit{Seamlessness}, we propose \seam{}, an automated method for estimating seamlessness between PM and RM. We validate its effectiveness through two experimental settings: data selection and augmentation.}
\label{teaser}
\end{figure}

Our study begins with the observation of a \textit{saturation phenomenon} in the RLHF process (\cref{saturation}): beyond a certain threshold, improvements in the quality of the RM and PM do not translate into increased RLHF performance (\autoref{teaser}). To understand this phenomenon, we explore whether the RM can assign appropriate scalar rewards to responses $r$ generated by the PM prompted by instruction $I$. This inquiry addresses the \textbf{seamlessness} between the RM and PM.
Although the RM performs well on standard preference benchmarks, it struggles to evaluate PM-generated responses effectively. This is demonstrated by a 35\% mismatch rate between reward scores and human preferences, indicating a significant, persistent discrepancy between the RM and PM as reflected in the reinforcement learning (RL) training data. This discrepancy does not diminish even as the PM and RM are individually optimized according to their respective evaluation paradigms, thus disrupting their seamlessness.
Remarkably, when we remove instructions from the RL dataset that contribute to this discrepancy and re-conduct RLHF, we observe an improvement in RLHF performance. This outcome suggests that enhancing the seamlessness between PM and RM benefits the overall RLHF process.

Based on these findings, we define the seamlessness between the PM and RM as detailed in \cref{estimate} and introduce an automated estimation method, \seam{}, available in three computational variants: \seam{}$_{\text{Adv}}$, \seam{}$_{\text{Contrast}}$, and \seam{}$_{\text{GPT}}$. Such methods remove the reliance on manual effort traditionally required for measuring seamlessness. Essentially, \seam{} evaluates the risk associated with each data sample when employed in RLHF processes, considering the specifics of the given PM and RM. Additionally, we give two experimental scenarios to demonstrate how \seam{} can be effectively utilized to improve the real-world RLHF process.
(1) Data Selection: We compute the \seam{} score for each sample and exclude those with low scores for RL training data selection. This strategy underscores a ``less is more'' phenomenon \citep{zhou2024lima}, whereby RLHF performance is enhanced when using this filtered dataset compared to the unfiltered dataset. Additionally, removing low-score samples helps mitigate the ``saturation phenomenon''.
(2) Model augmentation: During RLHF, we explore the PM and RM failure modes and subsequently strengthen them based on identified weaknesses. We calculate the \seam{} score for each data sample throughout the RL training phase. Samples exhibiting low \seam{} scores are then selected as targets for data augmentation to enhance the capabilities of the PM and RM specifically for these challenging samples. The results show that the \seam{} score effectively functions as a diagnostic metric within the RLHF framework. The primary contributions of this paper are three-fold:

\begin{itemize}[leftmargin=*]
\item We shift focus from the individual capacities of the reward model (RM) and policy model (PM) to explore their interplay and a noted saturation phenomenon in RM/PM quality. Our analysis identifies a discrepancy between RM and PM that cannot be resolved merely by scaling up.
\item We conceptualize the seamlessness between PM and RM and introduce \seam{}, an automatic estimation method that quantifies the seamlessness between PM and RM in a data-centric manner.
\item We empirically design two experimental scenarios to demonstrate how \seam{} can be leveraged to improve RLHF training: (1) Data selection and (2) Model augmentation. Our results validate the effectiveness of \seam{} under such scenarios.
\end{itemize}

\section{Related Work and Background}
\noindent \textbf{RLHF in Language Models.}
In earlier studies, reinforcement learning (RL) has been applied across various domains, such as machine translation \citep{sokolov2016bandit,kreutzer2018can,nguyen2017reinforcement}, dialogue generation \citep{li2016dialogue,yi2019towards,keneshloo2019deep}, and text generation \citep{li2018paraphrase,ziegler2019fine,shi2018toward,stiennon2020learning}, often employing modeling reward as automatic evaluation metrics like BLEU \citep{papineni2002bleu} or using simulated feedback \citep{nguyen2017reinforcement,keneshloo2019deep}. While integrating RL and language models has been extensively explored, significant advancements in RLHF with LLMs for general language tasks have only recently emerged \citep{ouyang2022training,touvron2023llama,achiam2023gpt,bai2023qwen,rafailov2024direct}. In RLHF, human feedback is collected to train a reward model, which then serves as a surrogate for human feedback during the training process, providing scalar evaluative feedback to the policy model (see detailed background of RLHF in \autoref{background}). In RLHF, RL algorithms (e.g., PPO \citep{schulman2017proximal}) are particularly suitable for training PM and RM.

\noindent \textbf{Reward Hacking.}
In RLHF, a critical issue closely related to our research is ``reward hacking'', as identified in prior studies~\citep{askell2021general,pan2021effects,skalse2022defining,shen2023trickle}. This phenomenon arises from \textbf{discrepancies between the reward model (RM) and actual human preferences} \citep{gao2023scaling,lambert2023alignment}. Although optimizing towards maximizing the rewards may initially appear beneficial, it ultimately leads the trained policy to exploit loopholes in the RM, securing high rewards without achieving the intended objectives. This degrades performance, complicates the selection of effective checkpoints, and may produce outputs that do not genuinely reflect human preferences \citep{singhal2023long}. Such misalignments increase tendencies towards sycophancy \citep{perez2023discovering}, reinforcing social biases \citep{santurkar2023whose,ziems2024can} and pose safety risks \citep{ngo2022alignment,carlini2024aligned,shen2024language}. A key distinction of our work is its focus on \textbf{the discrepancies between RM and PM}, which we term `seamlessness', as opposed to the traditional focus on discrepancies between reward models and human values.

\begin{figure}[t]
\centering
    \vspace{-2em}
    \includegraphics[width=\textwidth]{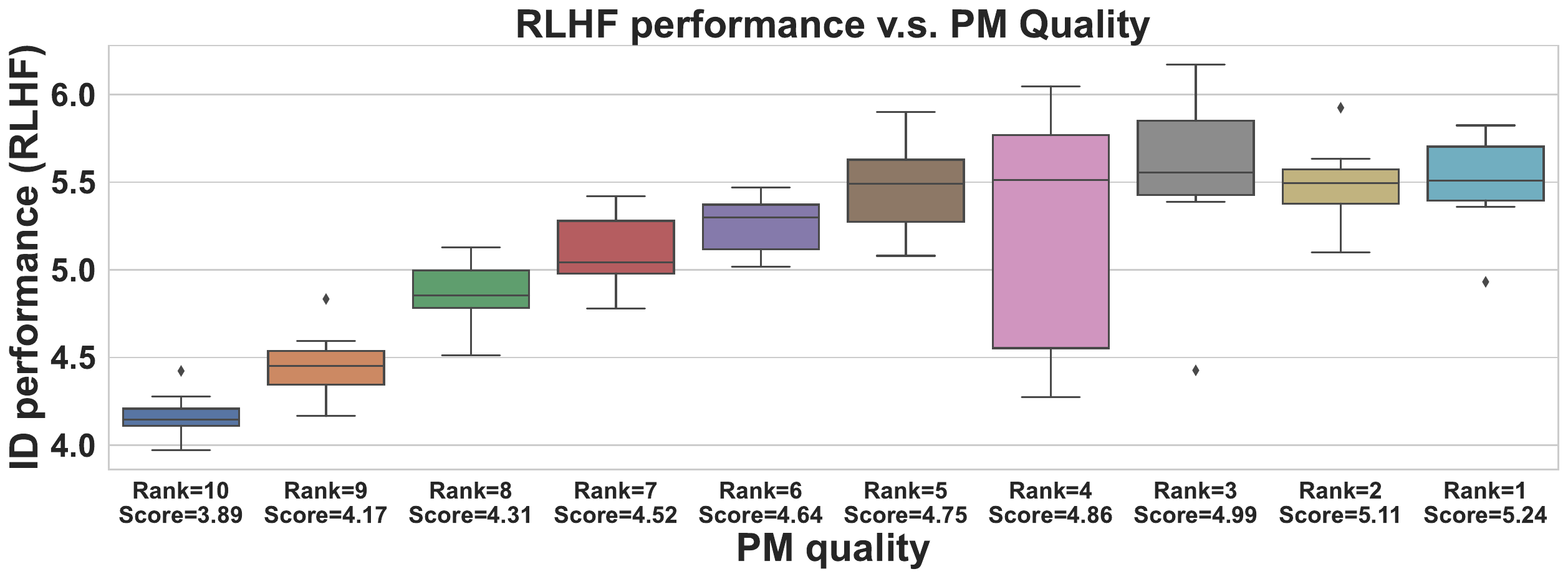}
    \includegraphics[width=\textwidth]{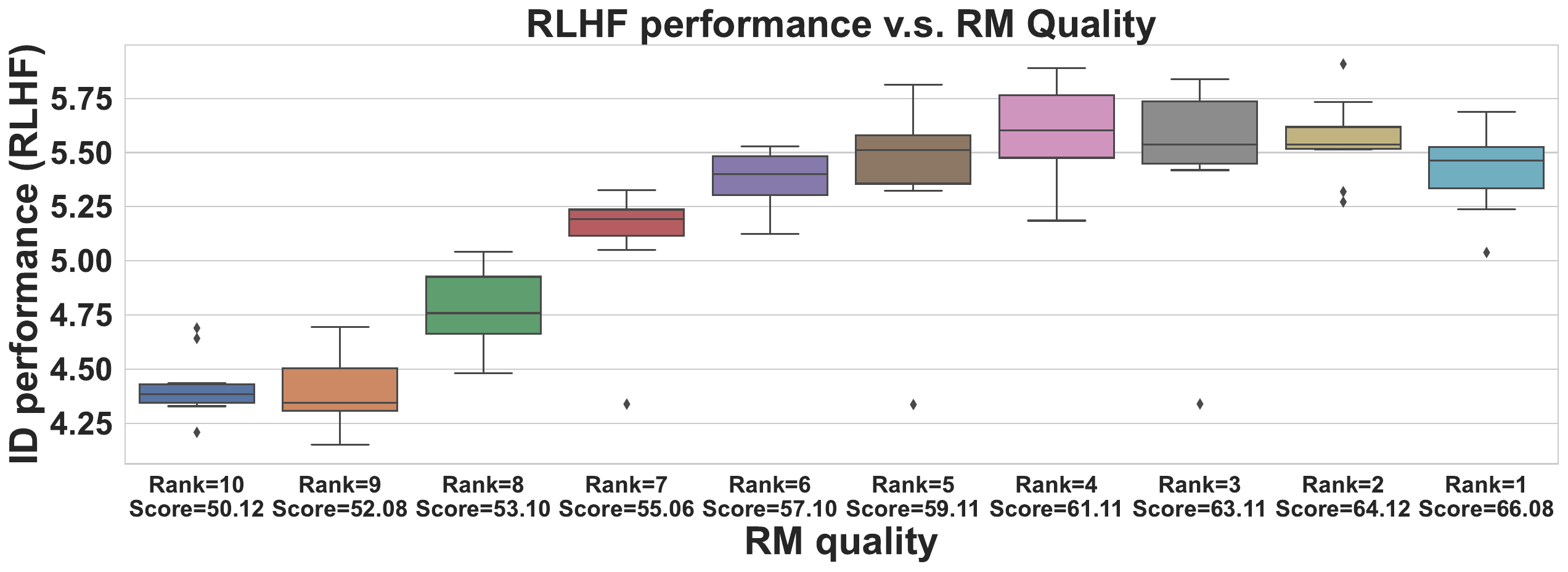}
    \vspace{-1em}
\caption{We examine the relation between the RLHF performance and the quality of PMs and RMs, measured by $\mathcal{P}_{in}$ and $\mathcal{A}_{in}$, respectively. We can see a ``saturation phenomeno'': the continual improvements of RM/PM do not translate into RLHF improvements.}
\vspace{-1em}
\label{fig:id}
\end{figure}

\section{The Saturation Phenomenon Reflected in RLHF Quality}
\label{saturation}
In this section, we conduct experiments to investigate the relationship between the RLHF outcomes and the quality of PM/RM. 

\textbf{Experimental Setup.}
We follow the experimental configuration of StackLLaMa \citep{beeching2023stackllama} due to the proven success of its PPO and data settings for RLHF. Our framework employs the \texttt{LLaMa2-7B} model as the base model for both the reward and policy models. To explore the effects of the quality of RM and PM, we change the volume of training data, enabling us to produce a spectrum of model strengths for both PM and RM. We develop ten variants each for RMs and PMs. Each pairing of PM and RM is then subjected to the RLHF technique, resulting in hundreds of unique RLHF models. Further details on implementation and setup are provided in \autoref{detail}.


\textbf{Quality Metrics.}
We employ two metrics\footnote{We do not use the KL divergence between the outputs from the reference and policy models, as there is no clear correlation between model quality and such KL divergence.} to assess the quality of the PM and RM: $\mathcal{Q}_{PM}$ (PM quality) and $\mathcal{Q}_{RM}$ (RM quality). In our experiments on StackExchange, $\mathcal{Q}_{PM}$ measures how well the policy model generates answers to StackExchange questions. We use 1000 samples from the StackExchange test split, with responses generated by the LLM evaluated by GPT-4 on a scale from 1 (worst) to 10 (best), similar to the MT-Bench scale. On the other hand, $\mathcal{Q}_{RM}$ evaluates the accuracy of the reward model in predicting human preferences on the StackExchange preference benchmark test split. Additional details are provided in \autoref{detail}.


\textbf{Results.}
We show the correlation between the in-domain performance ($\mathcal{Q}_{PM}$ and $\mathcal{Q}_{RM}$) of RLHF models and the quality of RMs and PMs, as illustrated in \autoref{fig:id}. Our primary observation is that while the quality of RMs and PMs generally positively correlates with the in-domain performance of RLHF models, a \textit{saturation effect} is evident. Beyond a certain quality threshold, additional RM or PM quality improvements yield no further enhancements in the in-domain performance of RLHF models.



\begin{wrapfigure}{R}{0.5\textwidth}
\vspace{-5mm}
\centering
    \includegraphics[width=0.5\textwidth]{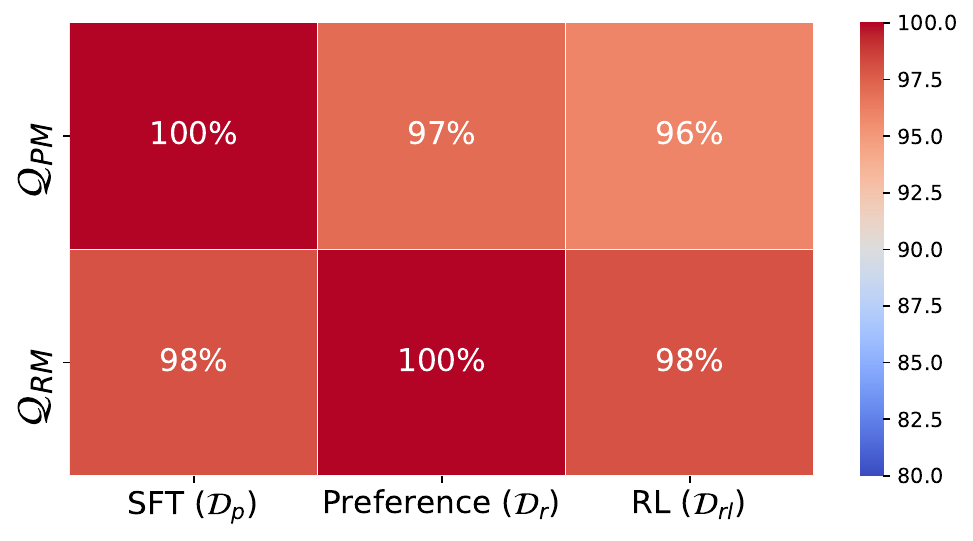}
\vspace{-5mm}
\caption{Cross-validation of PM and RM quality using different datasets(3 random seeds). The performance of RM and PM remains consistent across benchmarks. (e.g., on $\mathcal{D}{rl}$, the PM achieves 96\% of its performance on $\mathcal{D}_p$.)}
\vspace{-4mm}
\label{cross_validation}
\end{wrapfigure}

\section{Analyzing the Origin of Saturation Phenomenon}\label{analyze}
This section investigates the saturation phenomenon within RLHF, particularly from the perspective of potentially noisy supervision signals. The RLHF process comprises three primary stages: (1) policy modeling, (2) reward modeling, and (3) RL training. Initially, we conduct a sanity check on our PM and RM under our experimental settings to confirm their capacity for transferability across different data subsets. As shown in \cref{sanity}, the results indicate that both the PM and the RM exhibit adequate generalization. During the RL training stage, however, we observe that the RM struggles to effectively evaluate many of the responses generated by the PM (\cref{discrepancy}). By removing data that reflects this discrepancy between RM and PM, we find that RLHF performance improves (\cref{less}).

\subsection{A Sanity Check on PM and RM}\label{sanity}

We hypothesize that the observed saturation phenomenon may be due to the capacity of RM or PM can not be transferred to data used in other stages (e.g., the policy model can generate high-quality responses towards SFT instructions but fails to respond to the RL instructions).  Thus, we conducted a sanity check on both models to answer the following two questions: (1) Q1: whether the RM consistently distinguishes between better and worse responses as per the instructions used in SFT and RL training and (2) Q2: whether the PM sustains its generation quality with instructions from the RL dataset. We prepare the SFT dataset $\mathcal{D}_p$, the preference benchmark $\mathcal{D}_r$, and the RL dataset $\mathcal{D}_{rl}$. Specifically, the PM and RM were trained on the train splits of $\mathcal{D}_p$ and $\mathcal{D}_r$, respectively. We then employed cross-validation techniques to assess the PM's performance across the test split of the preference and RL datasets. Similarly, we tested the RM on the test split of the SFT and RL datasets. Experimental details are deferred to \autoref{detail}.

The results are shown in \autoref{cross_validation}. We trained five models each for the PM and RM, subsequently performing cross-validation. The key observation is that the performance of both PM and RM remains consistent across various in-domain datasets. This consistency indicates that PM and RM do not have significant generalization issues under our experimental setup. Besides, it also answers our two questions: (1) Given a well-trained PM that performs well on the evaluation set of $\mathcal{D}_p$, it can also respond with similar quality to the instructions in $\mathcal{D}_{rl}$; (2) Given a well-trained RM that performs well on the evaluation set of $\mathcal{D}_r$, it can also perform similarly well on distinguishing the golden and worse response in $\mathcal{D}_p$ and $\mathcal{D}_{rl}$.

\begin{wrapfigure}{R}{0.48\textwidth}
\vspace{-5mm}
\centering
    \includegraphics[width=0.48\textwidth]{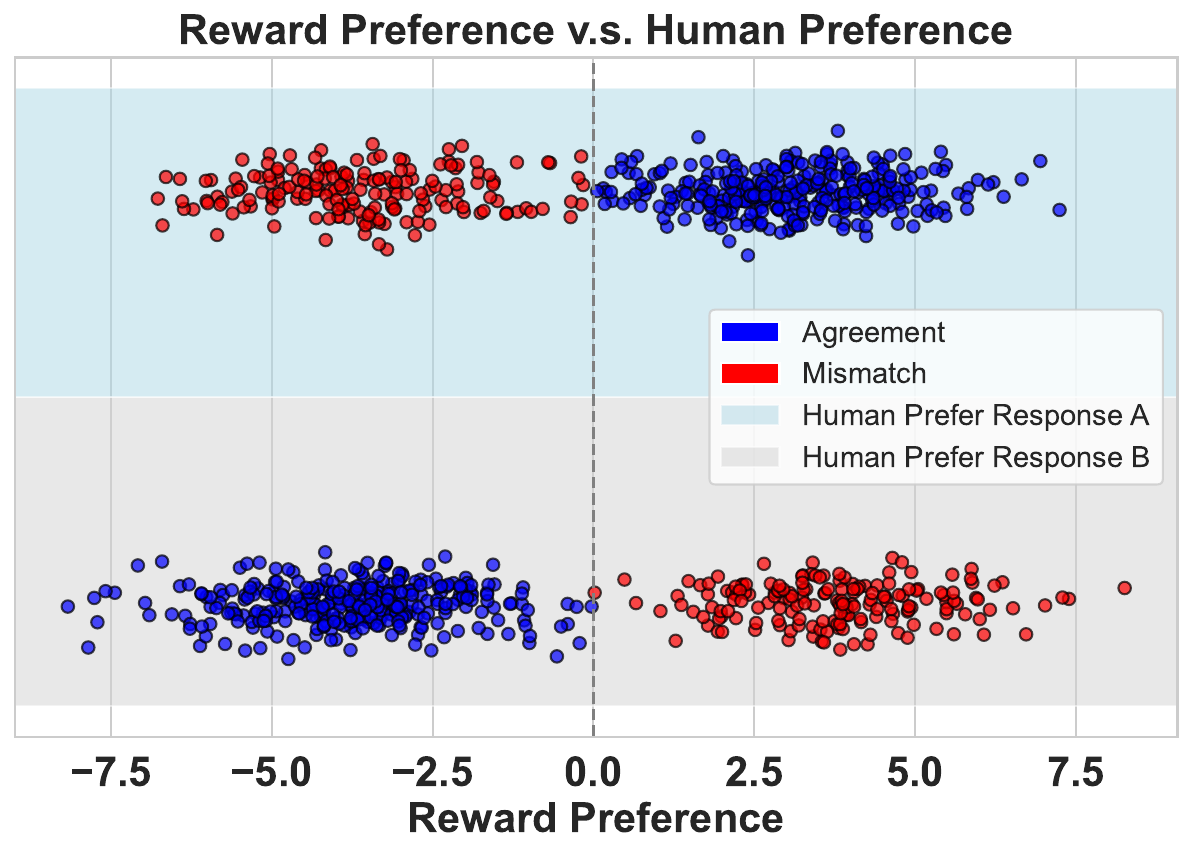}
\vspace{-5mm}
\caption{Agreement between reward and human preference is evaluated by comparing two responses (A and B) from two different policy models. The blue points indicate agreement between the reward and human preferences, while the red points represent mismatches. However, the results show that the RM fails to assign a proper score to the generation from PM.}
\vspace{-4mm}
\label{general_human}
\end{wrapfigure}

\subsection{Discrepancy between RM and PM during RL training}\label{discrepancy}
During the RL training stage, the PM is prompted by instructions from the RL dataset $\mathcal{D}_{rl}$ to generate responses $r_{i}$. The RM then evaluates these responses, which assigns reward scores to guide the RL training process. Our empirical analysis reveals two key findings (\autoref{cross_validation}), given a high-quality PM and RM: (1) the RM can effectively discriminate between golden and suboptimal responses of instructions within $\mathcal{D}_{rl}$, and (2) the PM can generate high-quality responses to instructions from $\mathcal{D}_{rl}$. Thus, we investigate the RM's capacity to evaluate the PM’s responses to $\mathcal{D}_{rl}$ since there might be a distribution shift between the responses generated from PM and those in the dataset.

Directly evaluating the RM capability to accurately assign scores to responses generated by the PM conditioned on an instruction $I_{i}$ has significant challenges since the standard reward modeling cast the preference regression problem into a classification problem. To address this, we employ a comparative analysis. We select two PMs of differing qualities (ranked 1 and 5 in previous experiments \cref{saturation}) and prompt each PM with instructions from the dataset $\mathcal{D}_{rl}$ (we sample a total of 1,000 instructions). We collect the responses and organize them into pairs for evaluation. Each pair of responses is evaluated by two methods: (1) human judgment and (2) RM evaluation using the rank 1 RM from \cref{saturation} to determine if even our best RM faces issues. To investigate the matching degree between RM and human preferences, we present pairs of responses (A and B) from the two PMs to human annotators without revealing the originating model. Human annotators are asked to annotate their preference between the two options. Similarly, we determine RM preferences based on their assigned reward scores.

The results, as shown in \autoref{general_human}, reveal a mismatch rate of approximately $40\%$, showing that the RM has some inability to accurately assign scores that reflect the true quality of responses generated by the PM. Also, we can observe a discrepancy between PM and RM - the RM can not well judge the quality of the responses generated from PM. This discrepancy can introduce noise into the RL training process, leading to the accumulation of incorrect gradients during RL optimization. Besides, we show that such discrepancies can not be resolved by scaling up the model (\autoref{discrepancy_scale}). Consequently, a natural strategy to enhance the RLHF process is removing instructions from $\mathcal{D}_{rl}$ that exhibit discrepancies between the RM and PM. This approach aims to reduce the noise in the RL training procedure, potentially improving overall model performance.

\begin{wrapfigure}{R}{0.45\textwidth}
\vspace{-7mm}
\centering
    \includegraphics[width=0.35\textwidth]{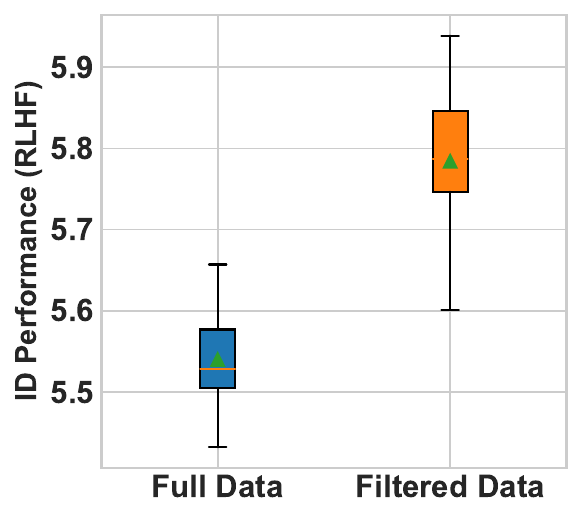}
\vspace{-4mm}
\caption{Compared to the RLHF performance of the full dataset, filter low-\seam{} data further improves RLHF (3 random seeds).}
\vspace{-5mm}
\label{case_comapre}
\end{wrapfigure}

\subsection{Less Can Be More: A Case Study of Data Selection for RL Training}\label{less}
Based on the insights from \cref{discrepancy}, we remove instructions that lead to discrepancies between the PM and RM. We then use this refined dataset for RL training and compare its performance against that achieved using the full $\mathcal{D}_{rl}$ dataset. As per the experimental settings described in \cref{discrepancy}, we employ both models at rank$=1$ for RL training. The results, presented in \autoref{case_comapre}, demonstrate a statistically significant improvement in RLHF performance (p<0.05) after removing data that causes discrepancies between the PM and RM. This case study illustrates a `less is more' phenomenon in RL training data: removing data that causes the discrepancy between PM and RM can enhance overall RLHF performance. However, this selective data filtering process is challenging to generalize due to its dependence on human annotation. Currently, there is no formal concept to characterize such data-driven discrepancies adequately. Consequently, we will discuss these in \cref{estimate}.

\section{\seam{}: An Automatic Estimation for Seamlessness}\label{estimate}
As shown in \cref{analyze}, removing data that leads to discrepancies between the PM and the RM improves RLHF performance. Currently, our approach depends on manual human assessments to determine the alignment between the PM and RM for specific datasets, a process that hinders full automation. This section first explores the concept of `seamlessness' in RL training data. Then, we propose \seam{}, an automated method designed to quantify the seamlessness of each data point, potentially enabling a more efficient and systematic tool to enhance RLHF training.

\subsection{Concept of the Seamlessness}
Generally, our concept of `seamlessness' is proportional to the PM likelihood of a data point that causes discrepancies between the policy and the reward model. Therefore, seamlessness includes not only the probability of misjudgment by the reward model but also the generative distribution of the policy model when conditioned on given data. The formal definition of seamlessness is provided in \autoref{definition1}. Considering that it is implausible to iterate the space of all responses $r$, we provide a discretization form for seamlessness in \autoref{discretization}.
 
\begin{definition}
    \label{definition1}
    (\textbf{Definition of Seamlessness}) Given an instruction ${I} \in \mathcal{D}_{rl}$, a reward model $\mathcal{R_{\theta}}$ and a policy model $\pi^{SFT}$. We denote the distribution of the response $r$ from $\pi^{SFT}$ as $P_{r}(\cdot |{I}, \pi^{SFT})$, we also denote the data distribution that hacks $\mathcal{R_{\theta}}$ as $P_{h}(\cdot | \mathcal{R_{\theta}})$, which means the data that leads to reward misjudgement. Then, the seamlessness of the instruction $I$ is defined as follows:
    \begin{equation}
        \mathcal{S}(I, {R_{\theta}}, \pi^{SFT}) = \int_{r \sim P_{h}} P_r\left(r \mid I, \pi^{S F T}\right)  \cdot \epsilon(r,R_{\theta}) \, dP_{h}
    \end{equation}
    where $\epsilon(r,R_{\theta})$ denotes the magnitude of RM misjudgement.
\end{definition}

Since the term defined in \autoref{definition1} is intractable, we propose \seam{}, an estimation for the seamlessness between RM and PM reflected through data. Following the notations in \autoref{definition1}, we define a sample set $\mathcal{X}$ that contains $N$ samples $r_{i} \sim P_{h}(\cdot | \mathcal{R_{\theta}})$ to represent the hacking distribution. Then, we present the discretization form of the seamlessness as follows:
\begin{equation}\label{discretization}
    \seam{}(I, {R_{\theta}}, \pi^{SFT}) = \sum_{r_{i} \in \mathcal{X}} P_r\left(r_{i} \mid I, \pi^{S F T}\right)\cdot \epsilon(r_{i},R_{\theta})
\end{equation}
In fact, our analyses in \cref{analyze} use a similar method to \autoref{discretization} to quantify the seamlessness between PM and RM. But under the formulation in \cref{analyze}, the $\epsilon(r_{i}, R_{\theta})$ refers to the mismatch degree between reward and human preferences, which inevitably incorporate the human efforts. 

\subsection{Automatic Estimation for Seamlessness}\label{automatic}
A significant practical challenge in our previous method of measuring seamlessness is the difficulty in automating the process. In this part, we introduce several automated estimation methods designed to quantify the seamlessness of data. Specifically, we propose three variants based on their corresponding designs to construct the sample set $\mathcal{X}$ (\autoref{discretization}): \seam{}$_{\text{Contrast}}$, \seam{}$_{\text{GPT}}$, \seam{}$_{\text{Adv}}$.

\paragraph{\seam{}$_{\text{Contrast}}$} In the \seam{}$_{\text{Contrast}}$ method, we implement the `Contrast Instruction' strategy \citep{shen2023trickle} to automatically construct the sample set $\mathcal{X}$. Specifically, for each instruction and its golden response pair $(I, r)$ in the dataset $\mathcal{D}_{rl}$, we retrieve 30 semantically relevant but distinct instructions $I^*$, along with their corresponding golden responses $r^*$, from a large SFT dataset (each pair in this dataset comprises an instruction and its golden response). We then use $r^*$ to form new pairs, assessing whether the reward model can effectively distinguish between the quality of the original pair $I \circ r$ and the newly constructed pair $I \circ r^*$. It is guaranteed that the quality of $I \circ r$ is superior to $I \circ r^*$, providing a reliable ground truth for evaluating RM performance. We define the magnitude of RM misjudgments, $\epsilon(r_i, R_{\theta})$, as follows:
\begin{equation}\label{magnitude}
    \epsilon(r_{i},R_{\theta}) = \text{max}\left\{R_{\theta}(I \circ r^{*}) - R_{\theta}(I \circ r), 0\right\}
\end{equation}
\paragraph{\seam{}$_{\text{GPT}}$} In the \seam{}$_{\text{GPT}}$ method, we use GPT-4 \citep{achiam2023gpt} to construct the sample set $\mathcal{X}$. Specifically, for each instruction and its golden response pair $(I, r)$ in the dataset $\mathcal{D}_{rl}$, we prompt GPT-4 to produce worse-quality responses $r^{*}$. Similarly, we use $r^*$ to form new pairs, assessing whether the reward model can effectively distinguish between the quality of the original pair $I \circ r$ and the newly constructed pair $I \circ r^*$. We reuse the magnitude defined in \autoref{magnitude}.

\paragraph{\seam{}$_{\text{Adv}}$} In the \seam{}$_{\text{Adv}}$ method, we use the adversarial attack to generate adversarial sentences that construct the sample set $\mathcal{X}$. Specifically, for each instruction and its golden response pair $(I, r)$ in the dataset $\mathcal{D}_{rl}$, we use adversarial attacks \citep{ren2019generating} to produce responses $r^{*}$ that hacks the reward model, such that $R_{\theta}(I \circ r^{*}) \textgreater R_{\theta}(I \circ r)$. Similarly, we follow the misjudgment term defined in \autoref{magnitude}.

\paragraph{Length penalty term} We introduce the operation to remove length bias. This operation targets the bias introduced by the length of response $r$, primarily affected by the exponential decrease in probability with increasing sequence length. To mitigate this, we implement a length normalization operation on the log probability of the response. This is formally represented as $\frac{\log P_r\left(r_{i} \mid I, \pi^{SFT}\right)}{\operatorname{len}(r_{i})}$, where $\log P_r(r_{i} \mid I, \pi^{SFT})$ denotes the logarithm of the probability that the policy model assigns to generating the response $r_i$ given the instruction $I$. It is also worth mentioning that, due to the logarithm, the value of \seam{} becomes negative.


\section{\seam{} for RL Training Data Selection}\label{seam_selection}
In this section, we employ three \seam{} variants as indicators to filter RL training data and evaluate the corresponding effectiveness.
\vspace{-1mm}
\subsection{Experimental setup}
Since this is a data-centric experiment, we follow the previous RLHF setup outlined in \autoref{detail}. For \seam{}$_{\text{Contrast}}$, we utilize SimCSE \citep{simcse2021} as the embedding model to retrieve the top 30 instructions from a StackExchange dataset containing over 1 million instruction-response pairs, with cosine similarity values in the interval $[0.8, 0.9]$. For \seam{}$_{\text{GPT}}$, we select \textsc{GPT-4-0613} to generate 30 lower-quality responses using the prompt shown in \autoref{prompt2}. For \seam{}$_{\text{Adv}}$, we employ TextAttack \citep{morris2020textattack} to perform adversarial attacks on the reward model. For each instruction, we generate 30 adversarial responses.

For the models, we reuse the policy model and reward model checkpoints from \cref{saturation} to calculate each \seam{} variant across the RL dataset. Subsequently, we filter out 20\% of the RL dataset based on the value of each \seam{} variant, respectively. We then compare the RLHF performance using the full and filtered datasets based on the evaluation paradigm used in \cref{saturation}. Specifically, we add a baseline (\textbf{LLaMa}) that uses the perplexity computed by LLaMa2-7B and filters the high perplexity data.

\begin{figure}[t]
\centering
    \includegraphics[width=\textwidth]{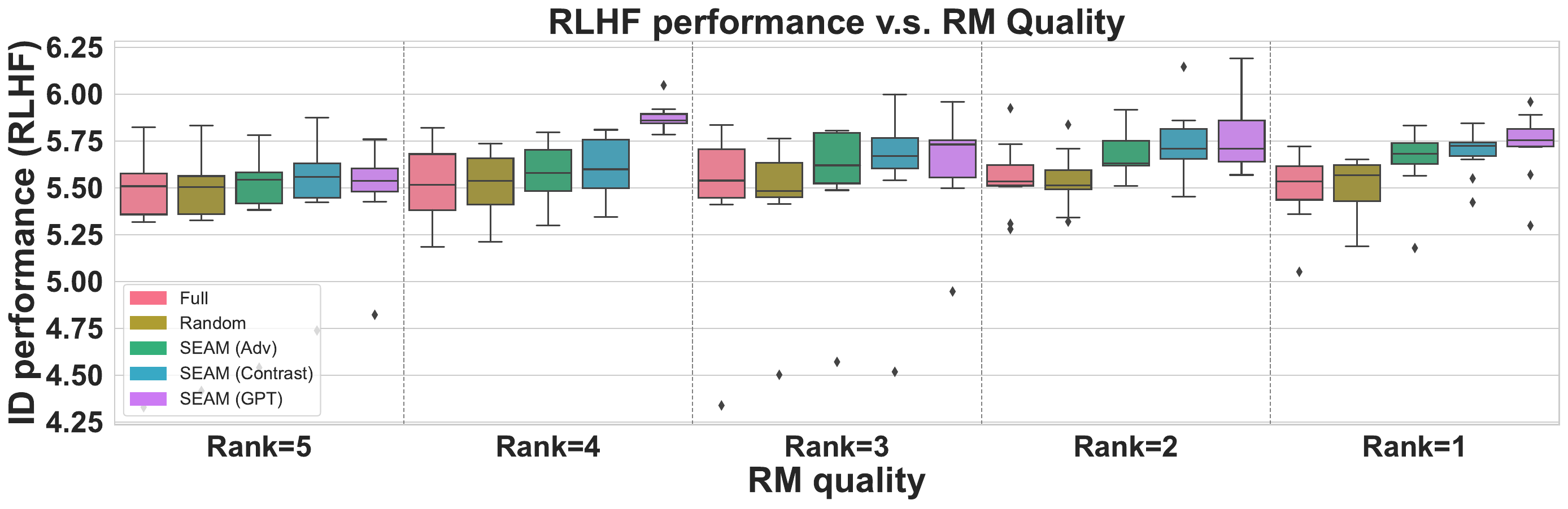}
    \includegraphics[width=\textwidth]{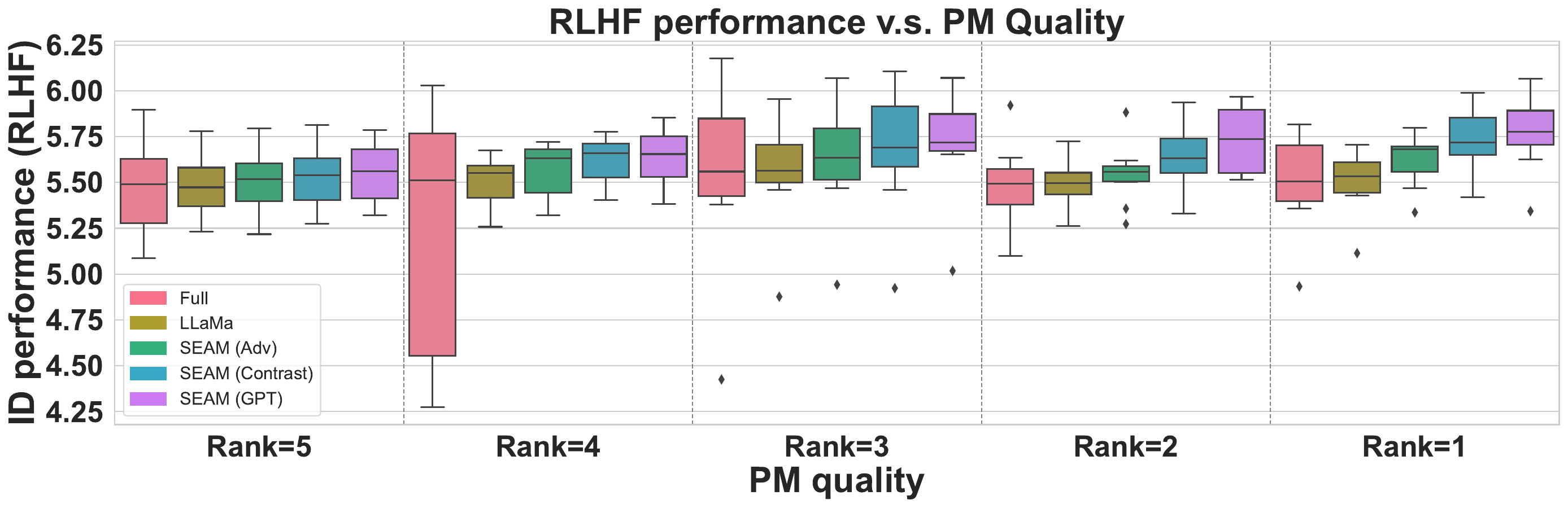}
\vspace{-2em}
\caption{RLHF performance when using \seam{} to filter 20\% of the RL dataset $\mathcal{D}_{rl}$. After filtering out the low-\seam{} data, we observe an improvement in RLHF performance compared to using the full $\mathcal{D}_{rl}$. The effectiveness of the three \seam{} variants is ranked as follows: \textbf{GPT} > \textbf{Contrast} > \textbf{Adv}. Specifically, we also observe that randomly removing 20\% RL data does not bring statistically significant performance changes.}
\label{fig:exp1}
\end{figure}

\vspace{-1mm}
\subsection{Results}
\vspace{-1mm}
The results are presented in \autoref{fig:exp1}, showcasing performance based on the top-5 RMs and PMs, where the saturation phenomenon occurs (\cref{saturation}). The key observations are as follows:

(1) Training on \seam{}-filtered RL data further improves RLHF performance: Compared to RLHF on the full $\mathcal{D}_{rl}$, conducting RL training on the filtered $\mathcal{D}_{rl}$ enhances RLHF performance. This finding empirically validates that data with low \seam{} values negatively impacts the RL training stage in RLHF. Additionally, randomly removing the same amount of RL training data does not yield benefits, indicating that the effectiveness of \seam{} is not merely due to a reduction in data size.

(2) Training on \seam{}$_{\text{GPT}}$-filtered RL data alleviates the saturation phenomenon: We observe that as the quality of RM (PM) increases, conducting RLHF on the data filtered by \seam{}$_{\text{GPT}}$ continues to improve performance to a certain extent. Compared to the case of full data training, the saturation phenomenon is mitigated by filtering data with low \seam{}$_{\text{GPT}}$ values. 

In general, the performance of the three \seam{} variants is ranked as follows: \textbf{GPT} > \textbf{Contrast} > \textbf{Adv}. In this section, we analyze the limitations of these variants through case studies and a straightforward analysis. Under the setup in \autoref{discretization}, a low likelihood indicates that, given the instruction $I$, the PM is unlikely to generate the response $r^{*} \in \mathcal{X}$, leading to issues in estimating seamlessness.

\begin{wraptable}{R}{0.55\textwidth}\centering
\begin{tabular}{ccc}
\toprule
\textsc{\seam{}}& \textsc{Attack} & \textsc{Likelihood}   \\
\midrule
\seam{}$_{\text{GPT}}$ & - &-1.81\\ \hdashline
\seam{}$_{\text{Contrast}}$ & - &-3.07\\ \hdashline
\multirow{3}{*}{\seam{}$_{\text{Adv}}$} & GA \citep{wang2019natural} & -9.32\\
& BA \citep{li2020bert} & -9.17\\
& PWWS \citep{ren2019generating} & -9.87\\
\bottomrule
\end{tabular}
\caption{Per-sentence log-likelihood (with length penalty) from the top-ranked PM (rank 1) for sentences in the sample set $\mathcal{X}$ (\autoref{discretization}) computed using the three estimation variants of \seam{}. The sentences created by \seam{}$_{\text{Adv}}$ exhibit significantly lower likelihoods, indicating their unnaturalness.}
\label{likelihood}
\end{wraptable}

For \seam{}$_{\text{Adv}}$, we found that the adversarial sentences generated for estimating \seam{} have a much lower likelihood in the PM compared to the other two methods, as shown in \autoref{likelihood}. Compared to the other two variants, the sentences generated by \seam{}$_{\text{Adv}}$ are significantly less likely to be sampled from the PM. Although such adversarial sentences can consistently hack the RM, they do not represent the PM’s natural outputs, indicating a lack of representativeness. This is because adversarial attacks tend to introduce non-coherent perturbations to the response $r$, significantly impacting fluency. We present typical cases in \autoref{seam_implementation}. For \seam{}$_{\text{Contrast}}$, a similar low-likelihood problem exists, although it is less severe than with \seam{}$_{\text{Adv}}$.

\section{\seam{} for RLHF Model Augmentation}\label{seam_diagnosis}
This section demonstrates how \seam{} can augment models that target to increase seamlessness between PM and RM.

\subsection{Experimental Setup}
We maintain our previous RLHF setup and use the same implementation of \seam{}. The key difference between this experiment and the one in \cref{seam_selection} is that, after computing \seam{} for the RL dataset, we augment the PM and RM by adding the data augmented based on such low-\seam{} data points in $\mathcal{D}_{rl}$, rather than filtering them.

For each \seam{} variant, we select the lowest 20\% of the data based on their \seam{} scores and generate augmented data to enhance the RM and PM. Specifically, for each low-\seam{} instruction $I_{i}$ and its corresponding golden response $r_{i}$, we apply the 'Contrast Instruction' strategy \citep{shen2023trickle} to create five augmented data samples for each instruction-response pair. These samples are then added to the training set of the PM. Similarly, we use the same method for the RM to generate five augmented preference data samples for each low-\seam{} instruction $I_{i}$, which are incorporated into the RM's training data. We assess the RLHF performance using the augmented PM and RM. To ensure a fair comparison, we add two baselines: (1) \textit{Random}: we randomly select 20\% of $\mathcal{D}_{rl}$ and apply the same augmentation method. The RLHF performance of the PM and RM augmented by both \seam{} and random selection is then evaluated. (2) \textit{Full Aug}: For each data sample in $\mathcal{D}_{rl}$, we apply augmentation methods based on all of them, and add the augmented data to train PM and RM.

\begin{wrapfigure}{R}{0.4\textwidth}
\vspace{-5mm}
\centering
    \includegraphics[width=0.4\textwidth]{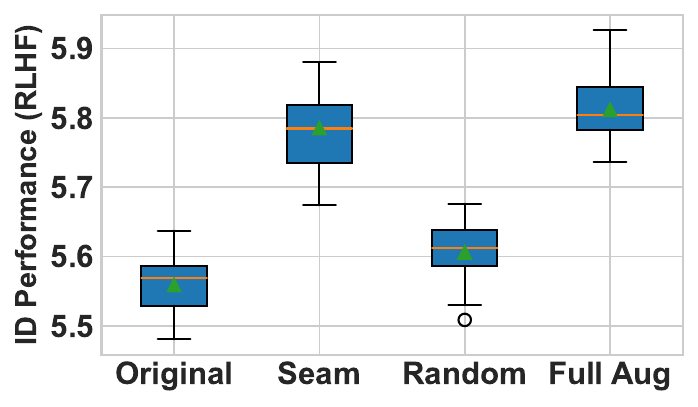}
\vspace{-4mm}
\caption{Performance comparison between model augmentation w/ and w/o \seam{}. `Original' means RLHF with no model augmentation.}
\vspace{-5mm}
\label{diagnosis}
\end{wrapfigure}

\subsection{Results}
As shown in \autoref{diagnosis}, the results illustrate the effectiveness of using \seam{} to guide model augmentation. Augmenting the PM and RM with data specifically selected by \seam{} demonstrates greater benefits than augmentations using randomly selected RL data and achieves comparable performance towards \textit{Full Aug}. 
This indicates that the RL data chosen by \seam{} is closely related to the weaknesses of the RM and PM combination during RLHF. Addressing these specific weaknesses through targeted data augmentation effectively improves the identified issues. 
Overall, this validates that \seam{} can serve as a signal to improve RM and PM in terms of their brittleness during RLHF.



\section{Limitations}
One limitation of our framework is that it is restricted to offline RLHF experiments rather than being tested in an online RLHF scenario. In online RLHF, the RM and PM are updated continuously based on real-time feedback from user interactions, offering a more dynamic and realistic setting. Despite this, we propose that \seam{} can still be effectively utilized by segmenting the online RLHF into a series of offline RLHF cycles. At the beginning of each cycle, the same analyses for data selection and model augmentation could be applied. This adaptation would allow us to extend the benefits of \seam{} to more practical, real-world applications. 
Another limitation of our framework is inherent in the \seam{} metric, which assesses the seamlessness of data only comparatively rather than absolutely. Consequently, while we can selectively filter portions of data (e.g., the lowest 20\%), we cannot establish a definitive threshold to categorize data as good or bad outright. However, to understand the impact of different filtering rates more thoroughly, we have conducted an analysis detailed in \autoref{filter}, where we see 20\% is a practical choice.

\section{Conclusion}
In this paper, we explored the concept of seamlessness between policy and reward models within Reinforcement Learning from Human Feedback (RLHF), uncovering significant discrepancies between the models as reflected in the data. We introduced \seam{}, an automated method to quantify this seamlessness, demonstrating its practical benefits for improving RLHF outcomes. Our findings emphasize the critical interplay between policy and reward models, contributing to a deeper understanding of RLHF dynamics. We hope our insights will guide future research toward developing more effective and nuanced RLHF strategies.

\bibliography{ref}
\bibliographystyle{apalike}

\newpage

\appendix

\section{Preliminaries: A three-stage paradigm for RLHF}
\label{background}
A RLHF practice includes three stages: policy modeling, reward modeling, and RL training, which involve three benchmarks: an SFT dataset $\mathcal{D}_p$, a preference benchmark $\mathcal{D}_r$, and an RL dataset $\mathcal{D}_{rl}$.
\paragraph{Policy model.}
Following the setup~\citep{ouyang2022training}, we obtain the policy model (PM) by supervised fine-tuning (SFT) the base version of LLM. Given an SFT dataset $\mathcal{D}_p$, each instance in the dataset consists of an instruction and its golden response. Then, we train the LLM on $\mathcal{D}_p$ with language modeling loss to obtain the PM: $\pi^{\mathrm{SFT}}$.

\paragraph{Reward model.}
Following the conventional setup~\citep{ouyang2022training}, we are given a dataset of human preferences $\mathcal{D}_r$. Each instance in this dataset ${(I_i, r_{i}^{+},r_{i}^{-})}$ is comprised of an instruction prompt $I_{i}$, a pair of responses $r_i^{+}, r_i^{-}$  where $r_i^{+}$ is preferred over $r_i^{-}$ by humans. On this labeled data, RM $\mathcal{R}_\theta$ is trained to assign a higher scalar reward to human-preferred $r_i^{+}$ over non-preferred $r_i^{-}$ in the context of $I_i$. 
This can be achieved by minimizing the ranking loss $\mathcal{L}$, where $\sigma$ is the sigmoid function and ${I}_{i}\circ r_{i}^{+}$ is the concatenation of $I_{i}$ and $r_{i}^{+}$.
\begin{equation}
\mathcal{L}(\theta)=-\E_{\left({I_{i}}, r_{i}^{+},r_{i}^{-}\right) \sim \mathcal{D}_h} \left[\log\left(\sigma \left(\mathcal{R}_\theta({I_{i}}\circ r_{i}^{+} \right)-\mathcal{R}_\theta \left({I_{i}}\circ r_{i}^{-}) \right) \right) \right].
\label{eq:rm}
\end{equation}

\paragraph{Reinforcement Learning.}
The last stage of RLHF is reinforcement learning. Specifically, a per-token KL penalty from the SFT model at each token is used to mitigate over-optimization of the reward model, and the value function is initialized from the RM. We maximize the following combined objective function $\mathcal{J}(\phi)$ in RL training based on PPO algorithm \citep{schulman2017proximal,ouyang2022training}, RL training dataset $\mathcal{D}_{rl}$ and pre-training dataset $\mathcal{D}_{\text {pre}}$:
\begin{equation}
 \mathcal{J}(\phi)=  \E_{(I,r) \sim \mathcal{D}_{\pi_\phi^{\mathrm{RL}}}}\left[\mathcal{R}_\theta(I\circ 
 r)-\beta \log \left(\pi_\phi^{\mathrm{RL}}(r \mid I) / \pi^{\mathrm{SFT}}(r\mid I)\right)\right] \nonumber 
\end{equation}
where $\pi_\phi^{\mathrm{RL}}$ is the learned RL policy parameterized by $\phi$ initialized from a pretrained supervised trained model $\pi^{\mathrm{SFT}}$. 
The first term encourages the policy $\pi_\phi^{\mathrm{RL}}$ to generate responses that have higher reward scores. 
The second term represents a per-token KL reward controlled by coefficient $\beta$ between $\pi_\phi^{\mathrm{RL}}$ and $\pi^{\mathrm{SFT}}$ to mitigate over-optimization toward the reward.

\section{The discrepancy does not vanish as scaling up}\label{discrepancy_scale}

\begin{table}[!h]\centering
\begin{tabular}{cccc}
\toprule
Model  & Match Rate  & PM performance & RM performance \\
\midrule
LLaMa2-7B & 60.5\% & 66.1 & 5.24\\
LLaMa2-13B & 60.7\% & 66.9 & 5.30\\
LLaMa2-70B & 60.4\% & 67.6 & 5.35\\
\bottomrule
\end{tabular}
\caption{The scaling tendency of our base model for training PM and RM, from 7B to 70B. We observe that the performance of PM and RM improves as the model scales up but find the match rate toward human preference remains nearly the same.}
\label{scale_model}
\end{table}

As demonstrated in \cref{discrepancy}, there is a notable discrepancy between the PM and RM: the RM fails to appropriately assign reward scores to responses generated by the PM. In this section, we explore the impact of scaling the base model on these discrepancies by reanalyzing the data discussed in \cref{discrepancy}. The findings, presented in \autoref{scale_model}, reveal that while the capacities of the PM and RM improve with an increase in the size of the base model (LLaMa2), the preference matching rate remains nearly consistent across different model scales. These results confirm that merely scaling up the model size does not address the underlying discrepancy between the RM and PM.

\section{Implementation details of RLHF}\label{detail}

\subsection{Training details}\label{detail_train}
• Standard fine-tuning (SFT): The base model chosen is LLaMa2-7B. We created 10 PMs of increasing quality by varying the training data amounts at 50, 100, 250, 500, 800, 1500, 2500, 5000, and 10000, plus a baseline pretrained model without SFT. 
The configuration employed includes the AdamW \citep{kingma2014adam} optimizer with a learning rate of 1e-4, 10 warmup steps, and training facilitated by LoRA. 
\\\\
• Reward model (RM): Training of the RM utilized the SFT model as the base model. Depending on the SFT model's quality rank, StackExchange pairwise preference data of subset 50, 100, 500, 2500, 5000, 10000, 20000, 50000, and 100000 were employed to train 9 RMs. With an additional pretrained model replaced with a randomly initialized classifier head, in total we create 10 RMs with increasing accuracy.
Training employed LoRA, with AdamW optimizer and learning rate 2e-5.
\\\\
• Reinforcement learning with PPO: 
PPO is used for each PM-RM pairing, generating hundreds of unique RLHF models. The RL prompts are from the StackExchange question dataset and remain consistent across all RLHF implementations. 
The SFT model served as the reference model, utilizing the reward scores from the RM as supervision. All PPO training has the configuration of LoRA with a learning rate of 1.4e-5, a batch size of 32, and 200 PPO steps.

\begin{prompt}\label{prompt1}
    (Prompt used in RLHF/PM evaluation)
    \vspace{2mm}
    
    [System]
    \vspace{2mm}
    
    Please act as an impartial judge and evaluate the quality of the response provided by an
    AI assistant to the user question displayed below. Your evaluation should consider factors
    such as the helpfulness, relevance, accuracy, depth, creativity, and level of detail of
    the response. Begin your evaluation by providing a short explanation. Be as objective as
    possible. After providing your explanation, please rate the response on a scale of 1 to 10
    by strictly following this format: "[[rating]]", for example: "Rating: [[5]]".
    \vspace{2mm}
    
    [Question]

    \{question\}
    \vspace{2mm}

    [The Start of Assistant’s Answer]
    
    \{answer\}
    
    [The End of Assistant’s Answer]
\end{prompt}

\subsection{Evaluation details}\label{detail_eval}
For the evaluation details, we detail the setup of the generator (i.e., PM and RLHF model) and classifier (i.e., RM), respectively. 

\begin{itemize}[leftmargin=*]
    \item Reward model: the reward model is evaluated on the corresponding test split of the preference benchmark based on accuracy (i.e., whether the RM can distinguish the better and worse response in the context of the given instruction.) 
    \item Policy and RLHF model: we follow the general principle of MT-Bench \citep{zheng2023judging}. Specifically, we use their instruction (\autoref{prompt1}) to prompt GPT-4 for measuring the quality of the responses from the policy and RLHF models. GPT-4 will assign a quality score, ranging from 0 to 10, to measure the quality of the response.
\end{itemize}

\subsection{Sanity check setup}\label{detail_sanity}
In the sanity check for the capacity of the RM and PM, our primary objective is to verify that both models maintain comparable performance across different stages of the training process. Specifically, we aim to ensure that: (1) the RM consistently distinguishes between better and worse responses as per the instructions used in SFT and RL training; (2) the PM sustains its generation quality with instructions from the RL training dataset.

To achieve this, we utilize the Stack-Exchange dataset's three segments (SFT, Preference, RL), dividing each into train, dev, and test splits. For the RM, the data distribution is 100,000/20,000/20,000, and for the PM, it is 20,000/2,000/2,000. We prepare the dataset in a format where each instruction is paired with a corresponding high-quality answer and a lower-quality candidate, ensuring the data's compatibility for training both the RM and PM. The training configurations adhere to the setup described in \autoref{detail}.

\section{Implementation details of \seam{}}\label{seam_implementation}
\subsection{Prompt used in \seam{}$_{\text{GPT}}$}
We use GPT-4 to generate worse-quality responses in \seam{}$_{\text{GPT}}$, based on the prompt detailed in \autoref{prompt2}.

\begin{prompt}\label{prompt2}
    (Prompt used in \seam{}$_{\text{GPT}}$)
    \vspace{2mm}
    
    [System]
    \vspace{2mm}
    
    Using the question and its correct answer provided below, generate 30 distinct answers that are of lower quality. Each response should include one or more of the following characteristics: factual inaccuracies, misunderstandings of the core question, irrelevant information, or grammatical errors. The answers should vary in their mistakes to cover a range of common errors seen in similar topics. Format the responses as separate paragraphs for each answer.
    \vspace{2mm}
    
    [Question]

    \{question\}
    \vspace{2mm}

    [Answer]

    \{answer\}
    \vspace{2mm}
    
    [The Start of Assistant’s Answer]
    
    \{answer\}
    
    [The End of Assistant’s Answer]
\end{prompt}

\subsection{Cases of \seam{}$_{\text{Adv}}$}
We employed several adversarial attack strategies to challenge the integrity of the reward model (RM). Specifically, for each instruction along with its corresponding better response $r^{+}$ and worse response $r^{-}$, these adversarial attacks introduce a perturbation $\alpha$ to $r^{-}$. The goal is for $r^{-} + \alpha$ to receive a higher reward score than $r^{+}$, thereby compromising the RM. The attacks we utilized include GA \citep{wang2019natural}, Bert-Attack \citep{li2020bert}, PWWS \citep{ren2019generating}, KATG \citep{shen2022katg}, and TextFooler \citep{jin2020bert}. However, a common limitation of these methods is that they tend to produce sentences with extremely low likelihood according to the policy model. Below, we present some examples illustrating the discrepancies between the original responses and those generated by the adversarial attacks.

\subsection{Setup of \seam{}$_{\text{Contrast}}$}
Using a human preference dataset, we have divided it into training, development, and testing sets. The reward model is trained on the training set and ceases training once it attains optimal performance on the development set. Subsequently, it is evaluated on the test set. Our \contrast{} are built upon the test set in each benchmark. We establish a similarity threshold range to ensure the retrieved instruction differs from the original one ($[0.8,0.9]$). Only instructions falling within this similarity range are retrieved.

\subsection{Human evaluation}
Since we aim to compute the degree of match between the reward outputs and human preferences, we enlist multiple human annotators to assess the quality of responses to Stack Exchange questions. Each annotator is kept unaware of the model that generated the responses, and then they are asked to give the index of the response with better quality based on tools like search engines. Since the evaluation relates to Stack Exchange, each annotator has expertise in computer science.

\section{Extra Analysis of low-\seam{} data}\label{filter}
\subsection{The effects of the filtering rate}
\begin{figure}
\centering
    \includegraphics[width=0.6\textwidth]{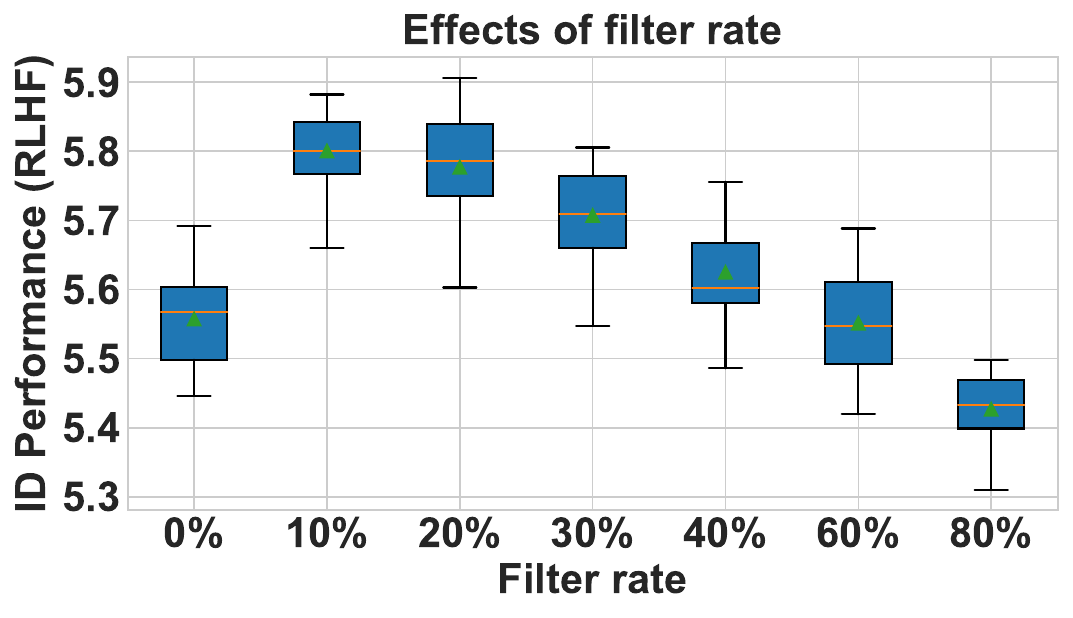}
\caption{The effects of filter rate in RL data selection.}
\label{filter_rate}
\end{figure}

We vary the filter rate as follows $\{10\%,20\%,30\%,40\%,60\%,80\%\}$, and re-conduct the experiments in \cref{seam_selection} with the rank 1 PM and RM. The results, as shown in \autoref{filter_rate}, demonstrate the relationship between the filter rate of data samples and the in-domain RLHF performance across various thresholds. Notably, increasing the filter rate initially enhances RLHF performance, with a peak observed at approximately 40\%. Beyond this threshold, further increases in the filter rate result in a gradual decline in performance. This trend indicates an optimal range for filtering out low-seam score samples to maximize RLHF effectiveness, thereby illustrating the critical trade-off between data quantity and quality. Based on this observation, we set the filtering rate as $20\%$.

\subsection{The overlap rate between low-\seam{} data on different combinations.}
Following the previous setup, we examine the overlap rate of the 20\% low-\seam{} data across three model combinations: (1) rank 5 PM with rank 5 RM, (2) rank 3 PM with rank 3 RM, and (3) rank 1 PM with rank 1 RM. We aim to assess whether the low-\seam{} data varies significantly among different model pairings. The results, illustrated in \autoref{overlap_rate}, reveal that the overlap rate between model combinations is generally high, exceeding 60\%. Notably, the overlap rate increases as the differences between the models decrease.

\begin{table}[!h]\centering
\begin{tabular}{cccc}\toprule
Model Combo & rank = 1 & rank = 3 & rank = 5 \\ \midrule
rank = 1    & -        &          &          \\
rank = 3    & $72\%$     & -        &          \\
rank = 5    & $64\%$     & $69\%$     & -   \\ \bottomrule    
\end{tabular}
\caption{The overlap rate between the 20\% low-\seam{} data on different model combinations, where a rank of 1 denotes using the rank 1 PM and rank 1 RM in the combination.}
\label{overlap_rate}
\end{table}

\section{Broader Impact}\label{broader}
Improved Human Model Alignment: Integrating \seam{} into RLHF techniques enhances the alignment between machine outputs and human values, leading to AI systems that are more ethical and responsive to user needs. This improvement is critical for increasing trust and encouraging the adoption of AI technologies across diverse sectors.

Increased Efficiency and Accessibility: Refining interactions between policy and reward models optimizes the training processes and reduces the computational resources required, making AI technologies more accessible and affordable. This democratization of AI could lead to broader innovation and application.

Misuse in Content Generation: The enhancements that improve model quality and user experience can also be exploited to create misleading information. Such misuse may pose risks of spreading misinformation and violating privacy.

\newpage
\begin{figure}[h]
\centering
    \includegraphics[width=\textwidth]{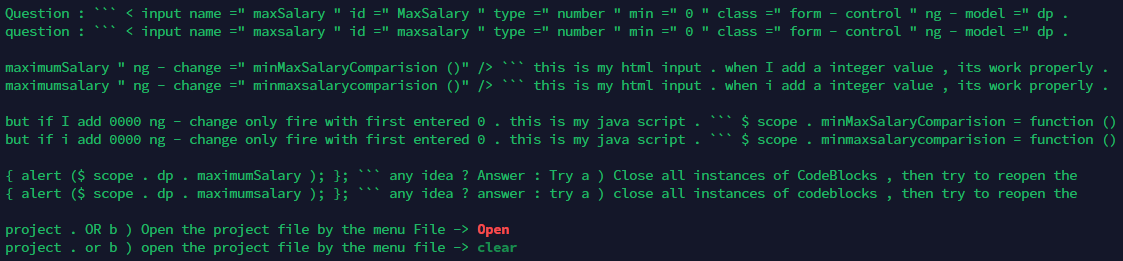}
    \vspace{3mm}
    \includegraphics[width=\textwidth]{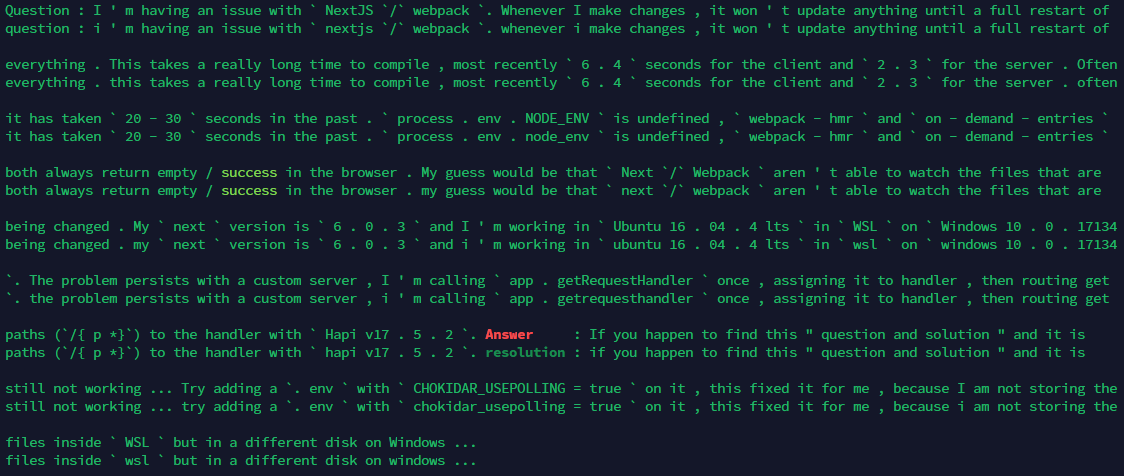}
    \vspace{3mm}
    \includegraphics[width=\textwidth]{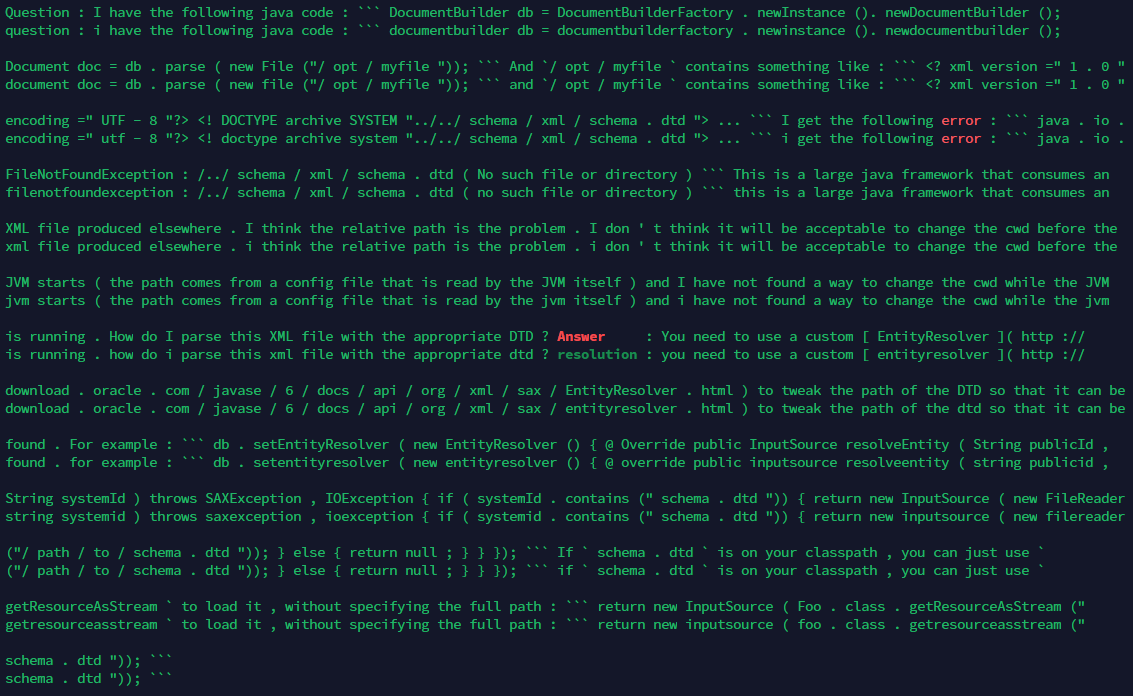}
\vspace{-4mm}
\caption{Case comparisons between the original and adversarial responses generated by text attacks. The differences are highlighted in \textcolor{red}{RED}.}
\label{fig:exp2}
\end{figure}

\end{document}